\documentclass[12pt]{article}
\usepackage{fancyhdr}
\usepackage{geometry}
\usepackage{wrapfig}
\usepackage[font = {footnotesize}]{caption}
\usepackage{graphicx} 
\usepackage[table]{xcolor}
\usepackage{colortbl}
\usepackage{multirow}
\usepackage{authblk}
\usepackage{natbib}
\usepackage{amsmath}
\usepackage{adjustbox}
\usepackage{orcidlink}

\usepackage{hyperref}
\usepackage{xcolor}
\hypersetup{
  colorlinks   = true, 
  urlcolor     = black, 
  linkcolor    = red, 
  citecolor   = blue 
}

\title{Crowdsourcing Without People: Modelling Clustering Algorithms as Experts}
\author{Jordyn E. A. Lorentz\orcidlink{0009-0004-8959-3873} and Katharine M. Clark\orcidlink{0000-0002-6162-2300}}
\date{\small Department of Mathematics \& Statistics, Trent University, Ontario, Canada}
\begin{document}

\maketitle


\begin{abstract}
    This paper introduces mixsemble, an ensemble method that adapts the Dawid-Skene model to aggregate predictions from multiple model-based clustering algorithms. Unlike traditional crowdsourcing, which relies on human labels, the framework models the outputs of clustering algorithms as noisy annotations. Experiments on both simulated and real-world datasets show that, although the mixsemble is not always the single top performer, it consistently approaches the best result and avoids poor outcomes. This robustness makes it a practical alternative when the true data structure is unknown, especially for non-expert users.
    
\end{abstract}


\section{Introduction} \label{Introduction}

Clustering is a widely used technique for uncovering patterns, or groupings, in unlabelled datasets. In model-based clustering, the data are assumed to arise from finite mixture model, whose density is given by
\begin{equation}\label{eq:mixmod}
	f(\mathbf{x} \mid \boldsymbol{\vartheta}) = \sum_{g=1}^{G} \pi_g f_g(\mathbf{x} \mid \boldsymbol{\theta}_g),
\end{equation}
where $\boldsymbol{\vartheta} = \{\pi_1, \dots, \pi_G, \boldsymbol{\theta}_1, \dots, \boldsymbol{\theta}_G\}$, $\pi_g>0$ with $\sum_{g=1}^G \pi_g = 1$, and $f_g(\mathbf{x} \mid \boldsymbol{\theta}_g)$ is the $g$th component density parameterized by $\boldsymbol{\theta}_g$.

The density, $f_g(\mathbf{x} \mid \boldsymbol{\theta}_g)$ should be chosen based on the structure of the clusters. However, the true distribution of the clusters is typically unknown before clustering. This makes it difficult to select the most appropriate algorithm. We propose an ensemble clustering approach, called mixsemble, based on the Dawid-Skene model \citep{Dawid79}, to address this uncertainty. Rather than selecting a single distribution, the proposed method uses the results of multiple clustering algorithms and combines them into one consensus clustering.

The result of this work is a clustering method that is more robust to uncertainty about a dataset's true structure. The proposed mixsemble method removes the burden of model selection from the user, which reduces the risk of making incorrect assumptions or relying too heavily on a single clustering algorithm. The findings of this paper provide valuable guidance in situations where the data distribution is unknown or difficult to model and where the use of a wide variety of component distributions provides a more balanced result.


\section{Literature Review} \label{Lit Review}

Ensemble clustering methods aim to aggregate cluster results into a single solution. The objective was first formalized by \cite{Strehl03}, who proposed graph partitioning methods. Other non-parametric approaches include co-association methods, \citep[e.g.,][]{Fred05,Huang18}, and voting methods \citep[e.g.,][]{zhou06}. In a parametric approach, \cite{topchy05} propose a model whereby clusterings are assumed to arise from a finite mixture model of multinomial distributions. Importantly, each algorithm's clustering is assumed to 
be independent and identically distributed. 

The Dawid-Skene model \citep{Dawid79} provides a method to aggregate labels from multiple observers. In particular, it does not assume that each observer is equally skilled. Using the EM algorithm, we obtain estimates for both the latent classes for each observation and the individual error-rate for each observer. This model has been applied in settings such as medical assessments \citep{Dawid79} and crowdsourced labelling \citep{Sinha18}.

Several extensions have been built on the basis of the Dawid-Skene model. Bayesian formulations \citep{Kim12} offer a solution for dependent data, while other work has introduced the Generative model of Labels, Abilities, and Difficulties (GLAD) method \citep{Whitehill09} or variational inference methods for graphical data \citep{Liu12}.

Our work proposes an extension to an ensemble method in model-based clustering based on the Dawid-Skene algorithm. Specifically, we adapt the algorithm to operate in the context of clustering results instead of responses from people. The methodology for this approach is outlined in the next section.


\section{Methodology} \label{Methodology}
The proposed methodology is a two-stage approach. First, the data are clustered separately using different distributions in model based clustering. Then, those algorithms are modelled as observers using the Dawid-Skene model. To do this, we require the following assumptions:
\begin{enumerate}
	\item True labels are drawn from a multinomial distribution with prior probabilities $\pi_g$ for $g\in [1,G]$
	\item Each clustering algorithm receives identical information with which to formulate its clustering.
	\item Given the true label, different algorithms' clusterings are assumed to be independent of each other.
\end{enumerate}
Assumption 1 is satisfied by assuming the data arise from a finite mixture model in \eqref{eq:mixmod}. Assumption 2 is satisfied by applying each algorithm separately. While Assumption 3 is not strictly satisfied, selecting algorithms with sufficiently different distributions can reduce correlations in their errors. This makes the independence approximation more reasonable. In this sense, the approach is analogous to the Na\"{i}ve Bayes classifier, which also relies on a conditional independence assumption that is rarely true in practice but often works well empirically.

 Let $\pi_g$ be the prior probability that an observation has true cluster $g$. Let $x_{ig}^{(k)}$ be an indicator variable with $x_{ig}^{(k)}=1 $ when algorithm $k$ clusters observation $i$ into cluster $g$, and $x_{ig}^{(k)}=0$ otherwise.  Let $z_{ig}$ be an indicator variable with $z_{ig}=1$ if observation $i$ truly belongs to cluster $g$, and $z_{ig}=0$ otherwise. Finally, let $\varepsilon_{gh}^{(k)}$ be an algorithm's error rate, i.e., the probability that algorithm $k$ classifies an observation truly in class $g$ into class $h$.  The likelihood of this model is given by:
\begin{equation} \label{eq:likelihood}
	\prod_{i=1}^N\biggl(\sum_{g=1}^G \pi_g \prod_{k=1}^K \prod_{h=1}^G (\varepsilon_{gh}^{(k)})^{x_{ih}^{(k)}} \biggr)
\end{equation} 

This model is fit using the EM algorithm, where in the E-step, the estimates for the $z_{ig}$'s are:
\begin{equation}  \label{eq:z}
	\hat{z}_{ig} = \frac{\pi_g\prod_{k=1}^K \prod_{h=1}^G  (\varepsilon_{gh}^{(k)})^{x_{ih}^{(k)}}}{\sum_{j=1}^G\pi_j\prod_{k=1}^K \prod_{h=1}^G (\varepsilon_{jh}^{(k)})^{x_{ih}^{(k)}}}.
\end{equation}
In the M-step, the parameters are estimated using their maximum likelihood estimates:
\begin{align}
  \hat{\varepsilon}_{gh}^{(k)} = \frac{\sum_i z_{ij}x_{ih}^{(k)}}{\sum_i\sum_h z_{ig}x_{ih}^{(k)}}, &   \qquad  \hat{\pi}_g = \frac{\sum_i z_{ig}}{N}.
\end{align}


The algorithm iterates between E- and M- steps until the log-likelihood converges. This provides estimates for the true classes, i.e. $\hat{z}_{ig}$, and the classification accuracy of each algorithm, i.e. $\hat\varepsilon_{gh}$.


\section{Experiments and Results} \label{Exp and Results}
In this section, we compare the performance of the proposed mixsemble approach to majority vote and different model-based methods individually. 
\subsection{Datasets}
We selected three simulated datasets and four widely used real datasets that show different data characteristics:
\begin{itemize}
    \item \textbf{\texttt{x2}} \citep{Pocuca24}: A dataset of 300 simulated observations across two variables for three clusters generated from a Gaussian distribution. 
    \item \textbf{\texttt{sx2}} \citep{Pocuca24}: A simulated dataset containing 2000 observations on two variables for two clusters generated bivariate variance-gamma distribution. 
    \item Manly \citep{Zhu23}: A simulated dataset containing 1000 observations across two variables for three clusters. Data are simulated using the inverse Manly transformation and the parameters in the example in the \texttt{ManlyMix} package for the \texttt{Manly.sim} function. 
    \item \textbf{\texttt{wine}} \citep{McNicholas25}: A dataset containing 178 observations on 27 properties for three types of wine.
    \item \textbf{\texttt{diabetes}} \citep{Scrucca23}: A dataset with observations for 145 diabetic patients on three variables for three classes of diabetes.
    \item \textbf{\texttt{iris}} \citep{Rcore24}: A dataset with 150 observations on four measurements for three species of iris flowers.
    \item \textbf{\texttt{seeds}}\citep{Zhu23}: A dataset with 210 observations across seven variables for three varieties of wheat kernels.
\end{itemize}

\subsection{Clustering Algorithms}
We compared the results of the following clustering algorithms:
\begin{itemize}
    \item {Generalized Hyperbolic Parsimonious Clustering Models} \citep[GH;][]{Browne15}
    \item {Gaussian Parsimonious Clustering Models} \citep[GPCM;][]{Celeux95}
    \item {Manly Mixture Modelling} \citep[ManlyMix;][]{Zhu18} 
	\item Majority vote \citep[e.g.,][]{zhou06}
    \item {mixsemble}-- the proposed method
\end{itemize}

Due to the way the datasets were generated, \texttt{sx2} is tailored to GH, \texttt{x2} to GPCM, and Manly to ManlyMix. Each algorithm is therefore expected to perform best on its corresponding simulated dataset. However, the purpose of the simulation study is to emulate a setting where the true structure is unknown, and thus direct matching between data generation and model is not assumed.

\subsection{Evaluation Framework}

We use the adjusted Rand index (ARI; \citeauthor{Hubert85}, \citeyear{Hubert85}) as our evaluation metric. The ARI measures agreement between partitions, in this case true and predicted classes. An ARI of 1 implies perfect agreement, 0 is expected when points are randomly partitioned, and negatives mean agreement is less than is expected with random chance.

\subsection{Comparative Analysis of Clustering Algorithms}

\begin{figure}[!htb]
	\centering
	\includegraphics[width=0.78\linewidth]{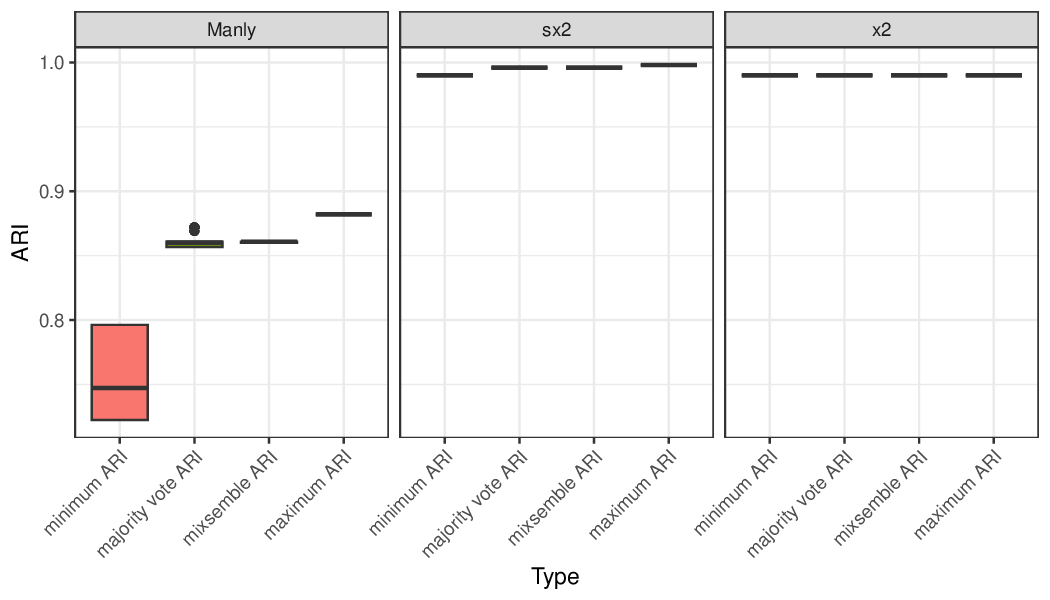}
	\includegraphics[width=\linewidth]{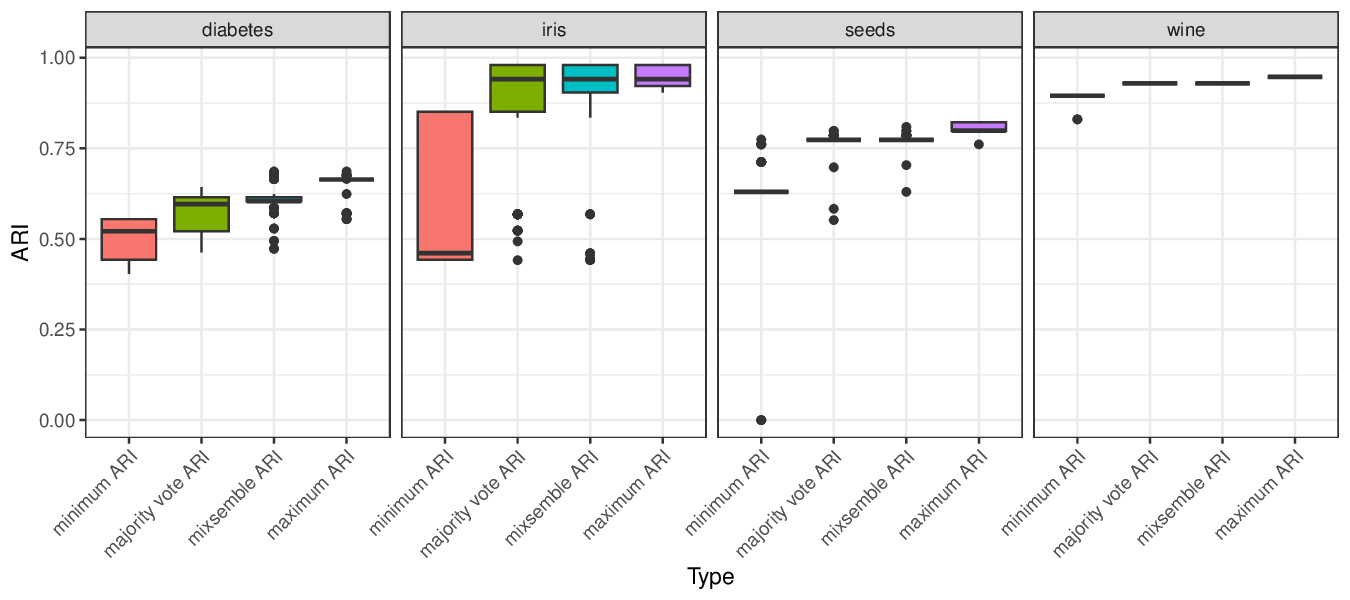}
	\caption{Performance comparison for ARI between mixsemble and the best/worst performing method}
	\label{fig:ARI_max_min}
\end{figure}

To assess clustering performance, we ran each algorithm on each of the previously stated datasets with 100 different random $k$-means initializations. Majority voting was performed by first aligning the clusters and then taking the most common classification. The mixsemble algorithm was run using the results from GH, GPCM, and ManlyMix as the  $x_{ig}^{(k)}$'s and initialized with $\hat{z}_{ig} = \frac{\sum_{k} x_{ig}^{(k)}}{\sum_{k}\sum_h x_{ih}^{(k)}}$. Table \ref{tab:ARI} presents a comparison of ARI for the clustering algorithms for each of the datasets. Table \ref{tab:ARI2} presents a comparison between the ensemble methods and the best- and worst-performing algorithms. Figure \ref{fig:ARI_max_min} displays the results in Table \ref{tab:ARI2} for each random initialization as a boxplot. 

\begin{table*}[htb!]
	\centering
		\begin{tabular}{lccc}
			\hline
			Dataset & GH & ManlyMix & GPCM  \\ 
			\hline
			x2 & \textbf{0.9900}(0) & \textbf{0.9900}(0) & \textbf{0.9900}(0) \\ 
			sx2 & 0.9945(0.0009) & 0.9900(0) & \textbf{0.9980}(0) \\ 
			Manly& 0.7552 (0.0320) & \textbf{0.8820} (0) & 0.7963 (0) \\ 
			wine & {0.9295(0)} & \textbf{0.9471}(0) & 0.8938 (0.0091) \\ 
			diabetes & 0.5271(0.0654) & 0.5506(0.0200) & \textbf{0.6454} (0.0547) \\ 
			iris & 0.7640(0.2373) & 0.7571(0.1683) & \textbf{0.9210} (0.0047)  \\ 
			seeds & {0.7754} (0.1371) & \textbf{0.7845}(0.0287) & 0.642(0.0363)  \\ 
			\hline
		\end{tabular}
	
	\caption{Mean ARI (with standard deviation in parentheses) computed over 100 random initializations. Bolded text indicates the best performing algorithm.}
	\label{tab:ARI}
\end{table*}

\begin{table}[htb!]
	\centering
\begin{tabular}{lcccc}
	\hline
	Dataset & Minimum ARI & Maximum ARI & Majority Vote & mixsemble \\
	\hline
	x2 & {0.9900}(0) & {0.9900}(0) & \textbf{0.9900}(0) & \textbf{0.9900}(0)   \\

	sx2 & 0.9900 (0)  & 0.9980 (0) & \textbf{0.9960} (0) & \textbf{0.9960} (0) \\

	Manly & 0.7543 (0.0308) & 0.8820 (0) & \textbf{0.8613} (0.0056) & 0.8605 (0.0004) \\

	wine & 0.8938 (0.0091) & 0.9471 (0) & \textbf{0.9295} (0) & \textbf{0.9295} (0)  \\

	diabetes & 0.5047 (0.0581) & 0.6547 (0.0304) & 0.5747 (0.0468) & \textbf{0.6121} (0.0422) \\

	iris & 0.6416 (0.2025) & 0.9504 (0.0260) & 0.8672 (0.1661) & \textbf{0.8691} (0.1804)  \\
	
	seeds & 0.6226 (0.1153) & 0.8066 (0.0113) & 0.7701 (0.0305) & \textbf{0.7732} (0.0172) \\
	\hline
\end{tabular}
\caption{Mean ARI (with standard deviation in parentheses) computed over 100 random initializations. The reported maximum and minimum ARI correspond to the highest and lowest values observed among the three algorithms in each run. The top performer between majority vote and mixsemble is bolded.}
\label{tab:ARI2}
\end{table}

These results show some interesting patterns:
\begin{enumerate}
     \item The performance of the clustering algorithms varies significantly across the datasets, suggesting that no one algorithm dominates for all types of data.
     \item The mixsemble method has a mean ARI that is the same or better than majority vote in all but one dataset. The mixsemble method has the best mean ARI in all real datasets. 
    \item The mixsemble method typically produces an ARI that is between  the worst- and best-performing methods. This ARI tends to be closer to the best-performing method than the worst-performing one. 
    \item The mixsemble approach tends to have lower variability than the majority vote approach. 
    \item \texttt{x2} and \texttt{sx2} both have very high ARI scores with no variability for all algorithms, indicating that clustering is very easy and choosing one algorithm over another is trivial.
\end{enumerate}


\section{Discussion} \label{Discussion}

This paper presented an extension of the Dawid-Skene algorithm to use clustering algorithms as observers. Our approach, called mixsemble, used the Dawid-Skene model \citep{Dawid79} to take a consensus measure of the results model-based clustering algorithms. We compared the performance of the proposed method with other clustering algorithms, including GH, GPCM, and ManlyMix. Our results show that while the proposed method does not always give the `best' result, it gives a result that is very similar to the best result and is often much better than the worst. In this sense, mixsemble acts like a risk-mitigation strategy-- it guards against poor outcomes by ensuring performance remains reliably strong, even when individual algorithms fail. Thus, the mixsemble algorithm is a robust alternative to relying on a single clustering method, particularly when the characteristics of the overall data structure are unknown. This may be especially useful to non-expert end users.

\vskip 0.2in
\bibliographystyle{chicago}
\bibliography{/Users/katclark/Local/Ensemble_method/TeX/MasterBibFile}

\end{document}